\documentclass[runningheads]{llncs}
\usepackage{graphicx}
\usepackage{comment}
\usepackage{amsmath,amssymb} % define this before the line numbering.
\usepackage{color}

\usepackage{amsthm}
\usepackage{color}  %- forbidden
\usepackage{dsfont}
\usepackage{mathtools}
\usepackage{amsmath}
\usepackage{amssymb}
\usepackage{algorithm}
\usepackage{algpseudocode}
\usepackage{multirow}
\usepackage{booktabs}
\usepackage{dblfloatfix}
\usepackage{subcaption}
\usepackage{url}
\usepackage{bm}
\usepackage{lineno}
\usepackage{hyperref} 
\usepackage{diagbox}
\usepackage{multirow}
\usepackage[bottom]{footmisc}

\begin{document}
\title{Pooling Regularized Graph Neural Network for fMRI Biomarker Analysis}
\author{Xiaoxiao Li$^{\star}$, 
%{Li, Xiaoxiao}
Yuan Zhou$^{\dagger}$,
%{Zhou, Yuan}
Nicha C. Dvornek$^{\dagger\star}$, 
%{Dvornek, Nicha C.}
Muhan Zhang$^{\lozenge}$, 
%{Zhang, Muhan}
Juntang Zhuang$^{\star}$, 
%{Zhuang, Juntang}
Pamela Ventola$^{\ddagger}$
%{Ventola, Pamela}
and James S. Duncan$^{\star\ast\dagger}$
%{Ducan, James S.}
}

% 1 {Li, Xiaoxiao}
% 2 {Zhou, Yuan}
% 3 {Dvornek, Nicha C.}
% 4 {Zhang, Muhan}
% 5 {Zhuang, Juntang}
% 6 {Ventola, Pamela}
% 7 {Duncan, James S.}

\authorrunning{X. Li et al.}


\institute{$^{\star}$ Biomedical Engineering, Yale University, New Haven, CT, USA \\
	$^{\ast}$ Electrical Engineering, Yale University, New Haven, CT, USA\\
	$^{\dagger}$Radiology \& Biomedical Imaging, Yale School of Medicine, New Haven, CT, USA\\
	$^{\ddagger}$ Child Study Center, Yale School of Medicine, New Haven, CT, USA\\
	$^{\lozenge}$ Facebook AI Research}
\maketitle              % typeset the header of the contribution
\begin{abstract}
Understanding how certain brain regions relate to a specific neurological disorder has been an important area of neuroimaging research. A promising approach to identify the salient regions is using Graph Neural Networks (GNNs), which can be used to analyze graph structured data, e.g. brain networks constructed by functional magnetic resonance imaging (fMRI). We propose an interpretable GNN framework with a novel salient region selection mechanism to determine neurological brain biomarkers associated with disorders. Specifically, we design novel regularized pooling layers that highlight salient regions of interests (ROIs) so that we can infer which ROIs are important to identify a certain disease based on the node pooling scores calculated by the pooling layers. Our proposed framework, Pooling Regularized-GNN (PR-GNN), %In contrast to feedforward neural networks (FNN) and convolutional neural networks (CNN) in traditional functional connectivity-based fMRI analysis methods, our proposed PR-GNN GNN only contains lower than  $1\%$ parameter. 
encourages reasonable ROI-selection and provides flexibility to preserve either individual- or group-level patterns. We apply the PR-GNN framework on a Biopoint Autism Spectral Disorder (ASD) fMRI dataset. We investigate different choices of the hyperparameters and show that PR-GNN outperforms baseline methods in terms of classification accuracy. The salient ROI detection results show high correspondence with the previous neuroimaging-derived biomarkers for ASD.

\end{abstract}

\section{Introduction}
\begin{figure}[t]
    \centering
    \includegraphics[width = 12cm]{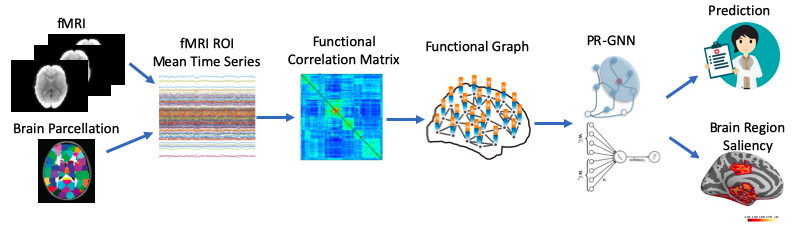}
    \caption{The overview of the pipeline. fMRI images are parcellated by atlas and transferred to graphs. Then, the graphs are sent to our proposed PR-GNN, which gives the prediction of specific tasks and jointly selects salient brain regions that are informative to the prediction task. }
    \label{fig:intro}
\end{figure}

\begin{figure}[t]
    \centering
    \includegraphics[width=12cm]{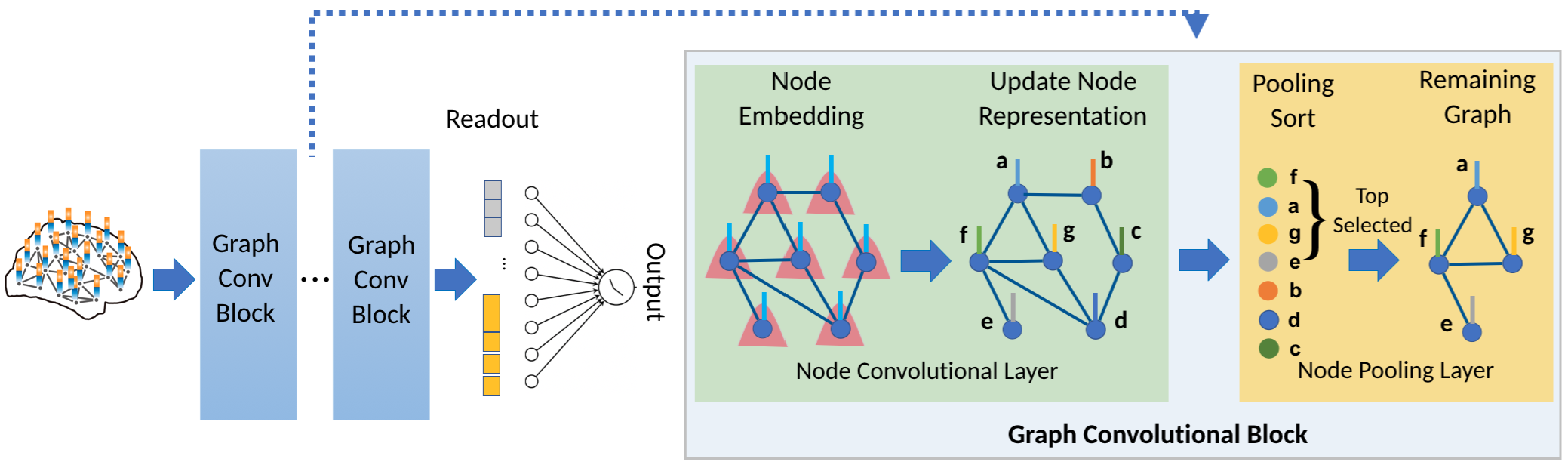}
    \caption{PR-GNN for brain graph classification and the details of its key component - Graph Convolutional Block. Each Graph Convolutional Block contains a node convolutional layer followed by a node pooling layer.  }
    \label{fig:gnn}
\end{figure}

Explaining the underlying roots of neurological disorders (i.e., what brain regions are associated with the disorder) has been a main goal in the field of neuroscience and medicine \cite{kaiser2010neural,goldani2014biomarkers,baker2014disruption,mcdade2018longitudinal}. %Neuroimaging techniques provide an objective view of measuring brain functions. 
Functional magnetic resonance imaging (fMRI), a non-invasive neuroimaging technique that measures neural activation, has been paramount in advancing our understanding of the functional organization of the brain \cite{worsley2002general,poldrack2009decoding,wang2019decoding}. The functional network of the brain can be modeled as a graph in which each node is a brain region and the edges represent the strength of the connection between those regions. 

The past few years have seen the growing prevalence of using graph neural networks (GNN) for graph classification \cite{hamilton2017inductive}.  Like pooling layers in convolutional neural networks (CNNs)  \cite{simonyan2014very,long2016unsupervised}, the pooling layer in GNNs is an important design to compress a large graph to a smaller one for lower dimensional feature extraction. Many node pooling strategies have been studied and can be divided into the following categories: 1) clustering-based pooling, which clusters nodes to a super node based on graph topology  \cite{defferrard2016convolutional,dhillon2007weighted,ying2018hierarchical} and 2) ranking-based pooling, which assigns each node a score and keeps the top ranked nodes \cite{gao2019graph,lee2019self}. Clustering-based pooling methods do not preserve node assignment mapping in the input graph domain, hence they are not inherently interpretable at the node level. For our purpose of interpreting node importance, we focus on ranking-based pooling methods. Currently, existing methods of this type \cite{gao2019graph,lee2019self} %may not be indictable (???) for finding salient nodes associated with graph classification tasks and their 
have the following key limitations when applying them to salient brain ROI analysis: 1) ranking scores for the discarded nodes and the remaining nodes may not be significantly distinguishable, which is not suitable for identifying salient and representative regional biomarkers,
and 2) the nodes in different graphs in the same group may be ranked totally differently (usually caused by overfitting), which is problematic when our objective is to find group-level biomarkers. To reach group-level analysis, such approaches typically require additional steps to summarize statistics (such as averaging). For these two-stage methods, if the results from the first stage are not reliable, significant errors can be induced in the second stage. 

%To overcome the limitations, 
To utilize GNN for fMRI learning and meet the need of group-level biomarker finding, we propose a pooling regularized GNN framework (PR-GNN) for classifying neurodisorder patients vs healthy control subjects and discovering disorder related biomarkers jointly. The overview of our methods is depicted in Fig.~\ref{fig:intro}. Our key contributions are:\\
$\bullet$ We formulate an end-to-end framework for fMRI prediction and biomarker (salient brain ROIs) interpretation.\\
$\bullet$ We propose novel regularization terms for ranking-based pooling methods to encourage more reasonable node selection and provide flexibility between individual-level and group-level interpretation in GNN.

\section{Graph Neural Network for Brain Network Analysis}

The architecture of our PR-GNN is shown in Fig. \ref{fig:gnn}. 
Below, we introduce the notation and the layers in PR-GNN. 
For simplicity, we focus on Graph Attention Convolution (GATConv) \cite{velivckovic2017graph,yang2019interpretable} as the node convolutional layer. For node pooling layers, we test two existing ranking based pooling methods: TopK pooling \cite{gao2019graph} and SAGE pooling \cite{lee2019self}.
\subsection{Notation and Problem Definition}
\label{sec:define}
\hfill \break
We first parcellate the brain into $N$ ROIs based on its T1 structural MRI.  We define ROIs as graph nodes $\mathcal{V} = \{v_1, \dots, v_N\}$. We define an undirected weighted graph as $G = (\mathcal{V},\mathcal{E})$, where $\mathcal{E}$ is the edge set, i.e., a collection of $(v_i,v_j)$ linking vertices  $v_i$ and $v_j$. $G$ has an associated node feature matrix $H=[\mathbf{h}_1, \dots, \mathbf{h}_{N}]^\top$, where $\mathbf{h}_i$ is the feature vector associated with node $v_i$. For every edge connecting two nodes, $(v_i, v_j) \in \mathcal{E}$, we have its strength $e_{ij}\in \mathbb{R}$. We also define $e_{ij}=0$ for $(v_i, v_j) \not\in \mathcal{E}$ and therefore the adjacency matrix $E=[e_{ij}]\in \mathbb{R}^{N \times N}$ is well defined.

\subsection{Graph Convolutional Block}
\subsubsection{Node Convolutional Layer}
To improve GATConv \cite{hamilton2017inductive}, we incorporate edge features in the brain graph as suggested by Gong $\&$ Cheng \cite{gong2019exploiting} and Yang et. al \cite{yang2019interpretable}. We define $\mathbf{h}_i^{(l)} \in \mathbb{R}^{d^{(l)}}$ as the feature for the $i^{th}$ node in the $l^{th}$ layer and $H^{(l)}=[\mathbf{h}_1^{(l)}, \dots, \mathbf{h}_{N^{(l)}}^{(l)}]^\top$, where $N^{(l)}$ is the number of nodes at the $l^{th}$ layer (the same for $E^{(l)}$). The propagation model for the forward-pass update of node representation is calculated as: 
\begin{equation} \label{eq:newembed}
\mathbf{h}^{(l+1)}_i = \phi_i^{\Theta^{(l)}}(H^{(l)}, E^{(l)}) =  \alpha_{i,i}\Theta^{(l)}\mathbf{h}_{i}^{(l)} +
        \sum_{j \in \mathcal{N}(i)} \alpha_{i,j}\Theta^{(l)}\mathbf{h}_{j}^{(l)},
\end{equation}
where the attention coefficients $\alpha_{ij}$ are computed as
\begin{equation}
    \alpha_{i,j}= \frac{\exp(\hat{\alpha}_{i,j})}{\sum_{k \in \mathcal{N}(i)\bigcup\{i\}}\exp(\hat{\alpha}_{i,k})},\; \hat{\alpha}_{i,j} =  e_{i,j}^{(l)}\text{ReLU}\big{(}(\mathbf{a}^{(l)})^\top[\Theta^{(l)} \mathbf{h}_i^{(l)} \| \Theta^{(l)} \mathbf{h}_{j}^{(l)}]\big{)},
\end{equation}
where $\mathcal{N}(i)$ denotes the set of indices of neighboring nodes of $v_i$, $\|$ denotes concatenation, $\Theta^{(l)}\!\in\! \mathbb{R}^{d^{(l+1)}\times d^{(l)}}$ and $\mathbf{a}^{(l)}\!\in\! \mathbb{R}^{2d^{(l+1)}} $ are model parameters.
\subsubsection{Node Pooling Layer}
The choices of keeping which nodes in TopK pooling and SAGE pooling are determined based on the node importance score $\mathbf{s}^{(l)} = [s_1^{(l)}, \dots, s_{N^{(l)}}^{(l)}]^\top$, which is calculated in two ways as follows:
\begin{equation}
    s_i^{(l)} = \begin{cases} \text{sigmoid}((\mathbf{h}_i^{(l)})^\top{\mathbf{w}^{(l)}}/\|\mathbf{w}^{(l)}\|), & \text{TopK pooling}\\ 
    \text{sigmoid}(\phi_i^{\boldsymbol{\theta}^{(l)}}(H^{(l)},{E}^{(l)})), & \text{SAGE pooling} \end{cases}
\end{equation}
where $\phi_i^{\boldsymbol{\theta}}$ 
%=[\phi_1,\dots,\phi_{N^{(l)}}]^\top$ 
is calculated in Eq. \eqref{eq:newembed} and $\mathbf{w}^{{(l)}}\in\mathbb{R}^{d^{(l)}}$ and $\boldsymbol{\theta}^{(l)} \in \mathbb{R}^{1 \times d^{(l)}}$ are model parameters. Note that $\boldsymbol{\theta}^{(l)}$ is different from $\Theta^{(l)}$ in Eq. \eqref{eq:newembed} such that the output of $\phi_i^{\boldsymbol{\theta}}$ is a scalar.

Given $\mathbf{s}^{(l)}$ the following equation roughly describes the pooling procedure:
\begin{equation} \label{eq:pool}
    \mathbf{idx} = \text{top}({\mathbf{s}}^{(l)},k^{(l)}), \quad 
    E^{(l+1)} = E^{(l)}_{\mathbf{idx},\mathbf{idx}}.
\end{equation}
The notation above finds the indices corresponding to the largest $k^{(l)}$ elements in score vector $\mathbf{s}^{(l)}$, and $(\cdot)_{\mathbf{i},\mathbf{j}}$ is an indexing operation which takes elements at row indices specified by ${\mathbf{i}}$ and column indices specified by ${\mathbf{j}}$. The nodes receiving lower scores will experience less feature retention. 

Lastly, we seek a “flattening” operation to translate graph information to a vector. Suppose the last layer is $L$, we use $\mathbf{z} = \text{mean}\, \{\mathbf{h}_i^{(L)}:i=1,\dots,N^{(L)}\}$, where mean operates elementwisely. Then $\mathbf{z}$ is sent to a multilayer perceptron (MLP) to give the final prediction.

\section{Proposed Regularizations}
\subsection{Distance Loss}
To overcome the limitation of existing methods that ranking scores for the discarded nodes and the remaining nodes may not be distinguishable, we propose two distance losses to encourage the difference. %we hope the top $k$ selected indicative ROIs should have significantly different scores than those of the unselected nodes. 
Before introducing them, we first rank the elements of the $m^{th}$ instance scores, ${\mathbf{s}}_m^{(l)}$, in a descending order, denote it as $\hat{\mathbf{s}}_m^{(l)}=[\hat{s}_{m,1}^{(l)},\dots,\hat{s}_{m,N^{(l)}}^{(l)}]^\top$, and denote its top $k^{(l)}$ elements as $a^{(l)}_{m,i}=\hat{s}_{m,i}^{(l)},i=1,\dots,k^{(l)}$, and the remaining elements as $b^{(l)}_{m,j}=\hat{s}_{m,j+k^{(l)}}^{(l)},j=1,\dots,N^{(l)}-k^{(l)}$. We apply two types of constraint to all the $M$ training instances. %First, we assume $k \leq N^{(l)}/2$, 

\subsubsection{MMD Loss}
Maximum mean discrepancy (MMD) loss \cite{gretton2012kernel,li2017mmd} was originally proposed in Generative adversarial nets (GANs) to quantify the difference of the scores between real and generated samples. In our application, we define MMD loss for the pooling layer as:
\begin{equation*}
\begin{split}
    L_{MMD}^{(l)} \! = \! -\frac{1}{M}\!\sum_{m=1}^M \! &\left[\frac{1}{(k^{(l)})^2}\sum_{i,j=1}^{k^{(l)}}  \kappa(a^{(l)}_{m,i},a^{(l)}_{m,j})\!+ \frac{1}{(N^{(l)}-k^{(l)})^2}\sum_{i,j=1}^{N^{(l)}-k^{(l)}} \kappa(b^{(l)}_{m,i},b^{(l)}_{m,j})\right.\\
    & \left. -\frac{2}{k^{(l)}(N^{(l)}-k^{(l)})}\!\sum_{i=1}^{k^{(l)}}\sum_{j=1}^{N^{(l)}-k^{(l)}}\!\kappa(a^{(l)}_{m,i},\!b^{(l)}_{m,j})\right],
\end{split}
\end{equation*}
where $\kappa(a,b)\! =\! \text{exp}(\parallel a\!-\!b \parallel^2)/\sigma$ is a Gaussian kernel and $\sigma$ is a scaling factor. 
\subsubsection{BCE Loss}
 Ideally, the scores for the selected nodes should be close to 1 and the scores for the unselected nodes should be close to 0. Binary cross entropy (BCE) loss is calculated as:
\begin{equation}
    L_{BCE}^{(l)} = -\frac{1}{M}\sum_{m=1}^M \frac{1}{N^{(l)}}\left[ \sum_{i=1}^{k^{(l)}}\log(a_{m,i}^{(l)})) + \sum_{i=1}^{N^{(l)}-k^{(l)}}\log(1-b_{m,i}^{(l)}) \right].
    \label{eq:dist1}
\end{equation}
 The effect of this constraint will be shown in Section \ref{sec:hyperpara}. 
\subsection{Group-level Consistency Loss}
\label{sec:gcl}
Note that $\mathbf{s}^{(l)}$ in Eq. \eqref{eq:pool} is computed from the input $H^{(l)}$. Therefore, for $H^{(l)}$ from different instances, the ranking of the entries of $\mathbf{s}^{(l)}$ can be very different. For our application, we want to find the common patterns/biomarkers for a certain neuro-prediction task. Thus, we add regularization to force the ${\mathbf{s}}^{(l)}$ vectors to be similar for different input instances in the first pooling layer, where the group-level biomarkers are extracted. We call the novel regularization group-level consistency (GLC) and only apply it to the first pooling layer, as the nodes in the following layers from different instances might be different. Suppose there are $M_c$ instances for class $c$ in a batch, where $c \in \{1,\dots, C\}$ and $C$ is the number of classes. We form the scoring matrix $S^{(1)}_c = [{\mathbf{s}}_{1,c}^{(1)},\dots,{\mathbf{s}}_{M_c,c}^{(1)}]^\top \in \mathbb{R}^{M^c \times N}$. The GLC loss can be expressed as:
\begin{equation} 
      L_{GLC}^c  = \frac{1}{M_c^2}\sum_{i=1}^{M_c}\sum_{j=1}^{M_c}\parallel{\mathbf{s}}_{i,c}^{(1)}-{\mathbf{s}}_{j,c}^{(1)}\parallel_2  
       = 2\text{Tr}((S_c^{(1)})^\top LS_c^{(1)}), \label{eq:con}
\end{equation}
where $L_c = D_c - W_c$, $W_c$ is a $M_c\times M_c$ matrix with all 1s, $D_c$ is a $M_c\times M_c$ diagonal matrix with $M_c$ as diagonal elements. We propose to use Euclidean distance for $\mathbf{s}_{i,c}$ and $\mathbf{s}_{j,c}$ due to 
the benefits of convexity and computational efficiency. 

Cross entropy loss $L_{ce}$ is used for the final prediction. 
Then, the final loss function is formed as:
\begin{equation}
    L_{total} = L_{ce} + \lambda_1 \sum_{l=1}^{L} L_{Dist}^{(l)} + \lambda_2 \sum_c^C L^c_{GLC},
    \label{eq:totalloss}
\end{equation}
where $\lambda$'s are tunable hyper-parameters, $l$ indicates the $l^{th}$ GNN block and $L$ is the total number of GNN blocks, $Dist$ is either MMD or BCE. 

\section{Experiments and Results}
\subsection{Data and Preprocessing}
We collected fMRI data from a group of 75 ASD children and 43 age and IQ-matched healthy controls (HC), acquired under the "biopoint" task \cite{Kaiser07122010}. The fMRI data was preprocessed following the pipeline in Yang et al.~\cite{yang2016brain}. The Desikan-Killiany \cite{desikan2006automated} atlas was used to parcellate brain images into 84 ROIs. The mean time series for each node was extracted from a random $1/3$ of voxels in the ROI by bootstrapping. In this way, we augmented the data 10 times. Edges were defined by top $10\%$ positive partial correlations to achieve sparse connections. If this led to isolated nodes, we added back the largest edge to each of them. For node attributes, we used Pearson correlation coefficient to node $1-84$. Pearson correlation and partial correlation are different measures of fMRI connectivity. We aggregate them by using one to build edge connections and the other to build node features. 

\subsection{Implementation Details}
%We trained and tested the algorithm on Pytorch using a NVIDIA Geforce GTX 1080Ti with 11GB GPU memory. 
The model architecture was implemented with 2 conv layers and 2 pooling layers as shown in Fig. \ref{fig:gnn}, with parameter $d^{(0)}=84, d^{(1)} =16, d^{(2)} =16$. We designed a 3-layer MLP (with 16, 8 and 2 neurons in each layer) that takes the flattened graph $\mathbf{z} \in \mathbb{R}^{16}$ as input and predicts ASD vs. HC. The pooling layer kept the top $50\%$ important nodes ($k^{(l)}=0.5N^{(l)}$). We will discuss the variation of $\lambda_1$ and $\lambda_2$ in Section \ref{sec:hyperpara}. We randomly split the data into five folds based on subjects, which means that the graphs from a single subject can only appear in either the training or test set. Four folds were used as training data, and the left-out fold was used for testing. Adam was used as the optimizer. We trained the model for 100 epochs with an initial learning rate of 0.001, annealed to half every 20 epochs. We set $\sigma=5$ in the MMD loss to match the same scale as BCE loss.
\subsection{Hyperparameter Discussion and Ablation Study}
\label{sec:hyperpara}
 
\begin{figure}[t]
    \centering
    \includegraphics[width=12cm]{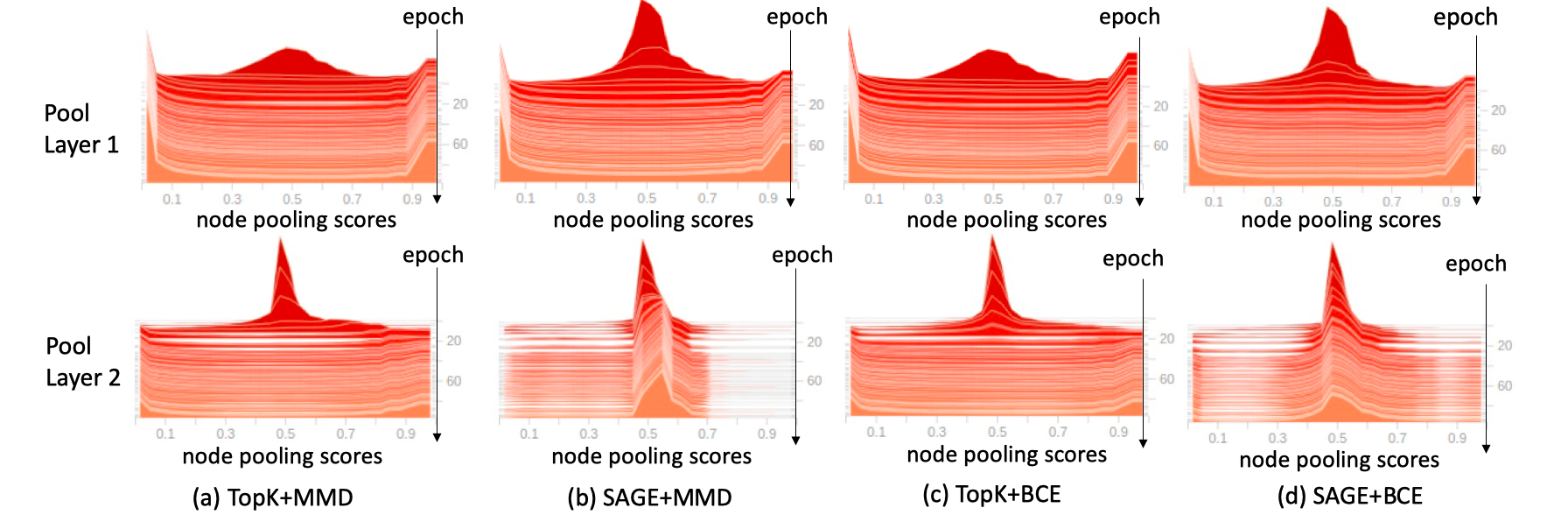}
    \caption{Distributions of node pooling scores over epochs (offset from far to near).}
    \label{fig:scores}
\end{figure}

We tuned the parameters $\lambda_1$ and $\lambda_2$ in the loss function Eq. \eqref{eq:totalloss} and showed the results in Table \ref{tab:model}. $\lambda_1$ encouraged more separable node importance scores for selected and unselected nodes after pooling. $\lambda_2$ controlled the similarity of the selected nodes for instances within the same class. A larger $\lambda_2$ moves toward group-level interpretation of biomarkers. We first performed an ablation study by comparing setting (0-0) and (0.1-0). Mean accuracies increased at least $3\%$ in TopK (1-$2\%$ in SAGE) with MMD or BCE loss. To demonstrate the effectiveness of $L_{Dist}$, we showed the distribution of node pooling scores of the two pooling layers in Fig. \ref{fig:scores} over epochs for %differnt model settings (denoted in Fig. \ref{fig:scores}), 
different combination of pooling functions and distance losses, 
with $\lambda_1=0.1$ and $\lambda_2=0$. In the early epochs, the scores centered around 0.5. Then the scores of the top $50\%$ important nodes moved to 1 and scores of unimportant nodes moved to 0 (less obvious for the second pooling layer using SAGE, which may explain why SAGE got lower accuracies than TopK). Hence, significantly higher scores were attributed to the selected important nodes in the pooling layer. %Small $\lambda_3$ would result in variant individual-specific patterns, while large $\lambda_3$ would force the model to learn common group-level patterns. %As task classification on HCP could achieve consistently high accuracy over the parameter variations, we only show the results on the Biopoint dataset in Table \ref{tab:table1} to better examine the effect of model variations.
Then, we investigated the effects of $\lambda_2$ on the accuracy by varying it from 0 to 1, 
with $\lambda_1$ fixed at 0.1. %As with other deep learning models, PR-GNN is overparameterized. 
Without $L_{GLC}$, the model was easier to overfit to the training set, while larger $L_{GLC}$ may result in underfitting to the training set. As the results in Table \ref{tab:model} show, the accuracy increased when $\lambda_2$ increased from 0 to 0.1  and the accuracy dropped if we increased $\lambda_2$ to 1 (except for TopK+MMD). For the following baseline comparison experiments, we set $\lambda_1$-$\lambda_2 $ to be $0.1$-$0.1$.

\begin{table}[t]
\centering
\caption{Model variations and hyperparameter ($\lambda_1$-$\lambda_2$) discussion.}
\label{tab:model}
\resizebox{0.8\textwidth}{!}{%
\begin{tabular}{@{}l|l|ccccc@{}} \toprule
\textbf{Loss} & \textbf{Pool} & 0-0 & 0.1-0 & 0.1-0.1 & 0.1-0.5 & 0.1-1 \\ \hline
\multirow{2}{*}{MMD} & TopK & 0.753(0.042) & 0.784(0.062) & 0.781(0.038) & 0.780(0.059) & 0.744(0.060) \\ 
 & SAGE & 0.751(0.022) & 0.770(0.039) & 0.771(0.051) & 0.773(0.047) & 0.751(0.050) \\ \hline
\multirow{2}{*}{BCE} & TopK & 0.750(0.046) & 0.779(0.053) & 0.797(0.051) & 0.789(0.066) & 0.762(0.044) \\ 
 & SAGE & 0.755(0.041) & 0.767(0.033) & 0.773(0.047) & 0.764(0.050) & 0.755(0.041) \\ \bottomrule
\end{tabular}%
}
\end{table}

\begin{table}[t]
\centering
\caption{
Comparison with different baseline models. }
\label{tab:comparison}
\resizebox{0.99\textwidth}{!}{%
\begin{tabular}{@{}l|cccccc@{}} \toprule
\multicolumn{1}{l|}{\diagbox[width=6em]{\textbf{Metric}}{\textbf{Model}}} & SVM & Random Forest & MLP & BrainNetCNN \cite{kawahara2017brainnetcnn} & Li et al. \cite{li2019graph} & \textbf{PR-GNN}$^*$\\ \hline
Acc & 0.686(0.111) & 0.723(0.020) & 0.727(0.047) & 0.781(0.044) & 0.753(0.033) & $\bm{0.797(0.051)}$\\
$\sharp$Par& 3k & 3k & 137k & 1438k &16k&$\bm{6k}$\\ 
\bottomrule
\end{tabular}
}
\small{Acc: Accuracy; $\sharp$Par: The number of trainable parameters; PR-GNN$^*$: TopK+BCE.}
\end{table}

\subsection{Comparison with Existing Models}
We compared our method with several brain connectome-based methods, including Random Forest (1000 trees), SVM (RBF kernel), and MLP (one 20 nodes hidden layer), a state-of-the-art CNN-based method, BrainNetCNN \cite{kawahara2017brainnetcnn} and a recent GNN method on fMRI \cite{li2019graph}, in terms of accuracy and number of parameters.  We used the parameter settings indicated in the original paper \cite{kawahara2017brainnetcnn}. The inputs and the architecture parameter setting (node conv, pooling and MLP layers) of the alternative GNN method were the same as PR-GNN. The inputs of BrainNetCNN were Pearson correlation matrices. The inputs of the other alternative methods were the flattened up-triangle of Pearson correlation matrices. Note that the inputs of GNN models contained both Pearson and partial correlations. For a fair comparison with the non-GNN models, we used Pearson correlations (node features) as their inputs, because Pearson correlations were the embedded features, while partial correlations (edge weights) only served as message passing filters in GNN models. The results are shown in Table \ref{tab:comparison}. Our PR-GNN outperformed alternative models. With regularization terms on the pooling function, PR-GNN achieved better accuracy than the recent GNN \cite{li2019graph}. Also, PR-GNN needs only $5\%$ parameters compared to the MLP and less than $1\%$  parameters compared to BrainNetCNN.
\subsection{Biomarker Interpretation}
\begin{figure}[t]
    \centering
    \includegraphics[width=11cm]{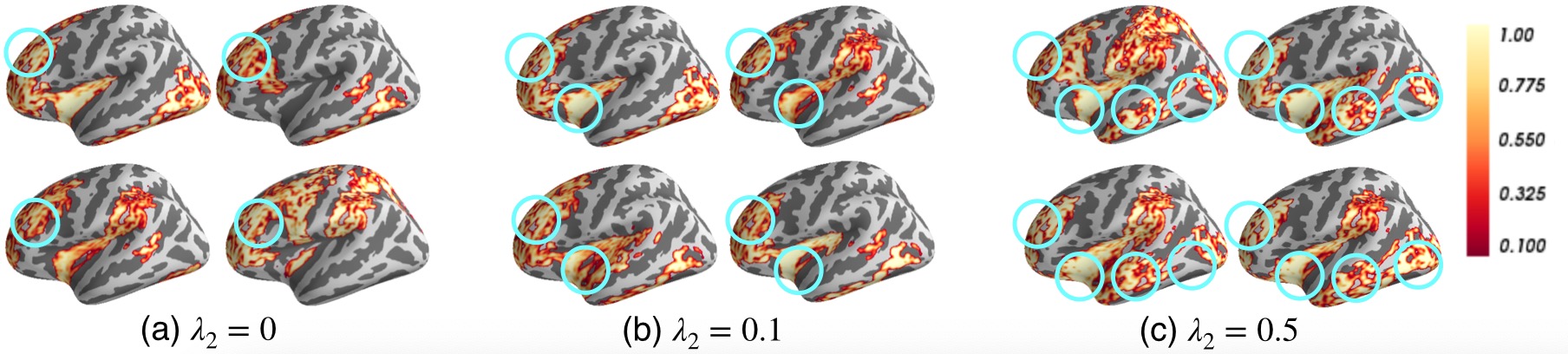}
    \caption{Selected salient ROIs (importance score indicated by yellow-red color) of four randomly selected ASD individuals with different weights $\lambda_2$ on GLC. The commonly detected salient ROIs across different individuals are circled in green.}
    \label{fig:lambda}
\end{figure}
Without losing generalizability, we investigated the selected salient ROIs using the model TopK+BCE ($\lambda_1=0.1$) with different levels of interpretation by tuning $\lambda_2$. As we discussed in Section \ref{sec:gcl}, large $\lambda_2$ led to group-level interpretation and small $\lambda_2$ led to individual-level interpretation. We varied $\lambda_2$ from 0-0.5. Without losing generalizability, we show the salient ROI detection results of four randomly selected ASD instances in Fig. \ref{fig:lambda}. We show the remaining 21 ROIs after the 2nd  pooling layer (with pooling ratio = 0.5, $25\%$ nodes left) and corresponding node pooling scores. As shown in Fig. \ref{fig:lambda}(a), when $\lambda_2 = 0$, we could rarely find any overlapped area among the instances. In Fig. \ref{fig:lambda}(b-c), we circled the large overlapped areas across the instances. By visually examining the salient ROIs, we found two overlapped areas in Fig. \ref{fig:lambda}(b) and four overlapped areas in Fig. \ref{fig:lambda}(c). By averaging the node importance scores (1st pooling layer) over all the instances, dorsal striatum, thalamus and frontal gyrus were the most salient ROIs associated with identifying ASD. These ROIs are related to the neurological functions of social communication, perception and execution \cite{schuetze2016morphological,hardan2006abnormal,bhanji2014social,press2012dissociable}, which are clearly deficient in ASD.

\section{Conclusion}
In this paper, we propose PR-GNN, an interpretable graph neural network for fMRI analysis. PR-GNN takes graphs built from fMRI as inputs, then outputs prediction results together with interpretation results. With the built-in interpretability, PR-GNN not only performs better on classification than alternative methods, but also detects salient brain regions for classification. The novel loss term gives us the flexibility to use this same method for individual-level biomarker analysis (small $\lambda_2$) and group-level biomarker analysis (large $\lambda_2$). We believe that this is the first work using a single model in fMRI study that fills the critical interpretation gap between individual- and group-level analysis. Our interpretation results reveal the salient ROIs to identify autistic disorders from healthy controls. Our method has a potential for understanding neurological disorders, and ultimately benefiting neuroimaging research. We will extend and validate our methods on larger benchmark datasets in future work.

\section*{Acknowledgements}
This research was supported in part by NIH grants [R01NS035193, R01MH100028].
%\clearpage
\bibliographystyle{ieeetr}
\bibliography{ref}
\end{document}